# An Automatic Question Usability Evaluation Toolkit

Steven Moore[1][0000-0002-5256-0339], Eamon Costello[2][0000-0002-2775-6006], Huy A. Nguyen[1][0000-0002-1227-6173], and John Stamper[1][0000-0002-2291-1468]

[1] Carnegie Mellon University, Pittsburgh PA 15213, USA
[2] Dublin City University, D09Y0A3 Dublin, Ireland
StevenJamesMoore@gmail.com

**Abstract.** Evaluating multiple-choice questions (MCQs) involves either labor-intensive human assessments or automated methods that prioritize readability, often overlooking deeper question design flaws. To address this issue, we introduce the Scalable Automatic Question Usability Evaluation Toolkit (SAQUET), an open-source tool that leverages the Item-Writing Flaws (IWF) rubric for a comprehensive and automated quality evaluation of MCQs. By harnessing the latest in large language models such as GPT-4, advanced word embeddings, and Transformers designed to analyze textual complexity, SAQUET effectively pinpoints and assesses a wide array of flaws in MCQs. We first demonstrate the discrepancy between commonly used automated evaluation metrics and the human assessment of MCQ quality. Then we evaluate SAQUET on a diverse dataset of MCQs across the five domains of Chemistry, Statistics, Computer Science, Humanities, and Healthcare, showing how it effectively distinguishes between flawed and flawless questions, providing a level of analysis beyond what is achievable with traditional metrics. With an accuracy rate of over 94% in detecting the presence of flaws identified by human evaluators, our findings emphasize the limitations of existing evaluation methods and showcase potential in improving the quality of educational assessments.

## 1 Introduction

Multiple-choice questions (MCQs) are the most commonly utilized assessment format across educational settings, spanning both traditional classroom environments and digital e-learning platforms [6]. Their versatility allows for assessing a broad spectrum of learning outcomes, ranging from simple recall to complex analytical skills, in many learning domains [17]. Besides offering grading efficiency and objectivity, MCQs enable the targeting of specific misconceptions through carefully crafted alternative answer options, known as distractors. However, the development of high-quality MCQs demands a rigorous approach to ensure reliability, validity, and fairness, essential for accurately measuring learners' knowledge and competencies [25].

Recent advances in natural language processing (NLP) have sought to alleviate the burden and time-consuming nature of MCQ authoring, enabling the rapid generation of questions at scale. These technologies facilitate the generation of hundreds of MCQs within minutes from sources such as document files or direct text requests [15]. Despite these advances, the rise in machine-generated MCQs has not uniformly translated to an improvement in quality. Machine-generated questions produced by state-of-the-art large language models (LLMs) often mirror the inaccuracies commonly found in human generated questions [7]. Such methods raise concerns regarding trust, authenticity, and diversity, potentially leading educators to be hesitant about adopting them without comprehensive evaluation [10].

Among the various MCQ evaluation techniques proposed in the literature, human judgment remains the gold standard, but typically faces challenges with subjectivity, time efficiency, and scalability [21]. Commonly used NLP metrics such as BLEU or METEOR, on the other hand, are much more efficient and scalable, but tend to focus on superficial features like readability and fail to align with human assessments or evaluate the *pedagogical* value of MCQs [12]. The effectiveness of MCQs are only as good as their design, requiring rigorous evaluation to ensure they serve as effective tools for assessing learning.

To address this gap, our research aims to establish a standardized and rigorous automated technique for MCQ evaluation. We begin by demonstrating the limitations of current NLP-based evaluation metrics, highlighting their lack of correlation with common errors found in MCQs. Then we introduce an automated evaluation technique, Scalable Automatic Question Usability Evaluation Toolkit (SAQUET), designed for comprehensive and standardized quality assessment of MCQs across multiple domains. Leveraging the 19 criteria of the Item-Writing Flaws (IWF) rubric [26], a proven and standardized instrument, SAQUET evaluates the structural and pedagogical quality of MCQs. We evaluate SAQUET across two datasets encompassing 271 MCQs from five diverse fields: Chemistry, Statistics, Computer Science, Humanities, and Healthcare.

The primary contributions of our work include: (1) providing empirical evidence on the inadequacy of prevalent MCQ quality evaluation metrics; (2) introducing SAQUET, an open-source tool capable of domain agnostic MCQ evaluation; and (3) compiling the most extensive and varied open dataset of MCQs annotated with IWF, providing opportunity for future research in educational assessment.

## 2   Related Work

### 2.1   Generating Multiple-Choice Questions

MCQs are widely recognized for their utility, but are susceptible to pattern matching and guessing [26]. However, with careful design, these issues can be mitigated, making well-crafted MCQs effective tools for evaluating a wide range of cognitive skills [15]. Crafting such high-quality questions is complex, with even the most advanced methods, including NLP and LLM based approaches, producing errors such as incorrect answers and nonsensical distractors [3]. For example, recent research showed that only 70% of machine generated MCQs for common subjects were deemed acceptable and clearly worded by human reviewers, with around 50% of the distractors being considered ineffective [22]. Furthermore, a 2023 survey revealed a hesitancy among educators to adopt these AI-powered tools, indicating a lack of trust in the generated MCQs [2].

Previous research has identified flaws in MCQs across various domains and teaching levels, including high-stakes standardized tests developed by psychometricians and domain experts [26]. These MCQs often find repeated use in test banks, practice sets, and training materials over the years. Consequently, there is a need for ongoing quality evaluation of these pre-existing questions, not just newly generated ones. This can complement analyses based on student performance data, such as those offered by Item Response Theory [1]. However, evaluating MCQs before their implementation is crucial to avoid exposing poorly designed questions to learners, which can impede their learning [23]. Crafting high-quality questions remains a significant challenge, and evaluating their quality poses an even greater one, demanding consistency, scalability, and consideration of the questions' application contexts.

## 2.2   Automated MCQ Quality Evaluation

Over the past decade, automated MCQ quality evaluation has relied on metrics such as BLEU, METEOR, and ROUGE [21]. These metrics primarily assess similarity to a gold standard without considering educational value or effectiveness in evaluating student knowledge [18]. While previous research states these "standardized" metrics facilitate comparison across studies, they involve numerous hyper-parameters that can vary by task and are often insufficiently reported, complicating precise comparisons and replications [16]. Moreover, prior work has demonstrated that these metrics do not sufficiently align with human evaluation [12, 27]. To align more closely with human evaluation while maintaining scalability, alternative automated approaches have explored metrics like perplexity, diversity, grammatical error, complexity, and answerability [24, 28]. These have been applied to both machine- and human generated questions, offering a broader evaluation that extends beyond mere readability to include aspects critical for educational assessments.

When evaluating MCQs, perplexity assesses a language model's ability to predict question and answer text based on its training data [4]. Lower scores suggest more coherent questions and answers with predictable language patterns, whereas higher scores indicate complexity or atypical text, suggesting the questions could be unclear or poorly structured. Diversity evaluates the range in vocabulary, structure, and content across generated texts, ensuring a variety of questions and answers and reducing repetition [13]. A higher diversity score indicates greater uniqueness among MCQs, avoiding repetitive phrases and templated patterns. Grammatical errors pinpoint grammar violations, such as incorrect verb tense or spelling, quantified for each MCQ.

Complexity is typically assessed through cognitive complexity, using Bloom's Revised Taxonomy to assign difficulty levels to MCQs based on the cognitive skills required to answer them [11]. Bloom's Revised Taxonomy categorizes cognitive skills ranging from recall (remembering) to higher-order skills (creating), with questions demanding higher-order thinking deemed more cognitively complex [7]. Answerability measures how accurately a question can be answered, using the provided context or common knowledge. Recently, LLMs such as GPT-4 have been used to automate this evaluation metric [24]. Specifically, the Prompting-based Metric on ANswerability (PMAN) strategy employs three prompts to evaluate a question's quality by how well an LLM can answer it, demonstrating that it aligns with human judgments [27].

## 2.3   Human MCQ Quality Evaluation

Despite the growth of automated methods for evaluating MCQ quality, human evaluation is still considered the benchmark for accuracy [11, 21]. However, this approach can be subjective, relying on vague metrics like "difficulty" or "acceptability" that are based on intuition [12]. Such evaluations are not only challenging to standardize but also difficult to scale, replicate accurately, and are time-intensive. A more objective alternative that has proven effective for over 15 years is the Item-Writing Flaws (IWF) rubric [26]. Comprising 19 criteria, the IWF rubric evaluates MCQs across any domain, focusing on pedagogical aspects beyond mere readability and surface-level features. It has been successfully applied to MCQs in diverse fields, including standardized medical exams, chemistry, and computer science MOOCs, demonstrating its utility in ensuring quality educational assessments [5, 19, 23].

## 3   Methods

### 3.1   Item-Writing Flaws (IWF) Rubric

In our study, we adopted the 19-criteria IWF rubric, a tool that has been validated and employed in prior research [5, 19, 23, 26]. The rubric is designed to be universally applicable across domains, encompassing both pedagogical considerations and factors related to human test-taking abilities. Unlike traditional metrics that primarily assess readability, the IWF rubric includes criteria that address a broader range of question quality aspects, such as unintentional hints, cues, and modality. Table 1 outlines each of the 19 criteria, providing guidance on avoiding specific flaws and ensuring adherence to the rubric's standards. Previous research indicates an MCQ with zero or one IWF can generally be considered acceptable for use, particularly in contexts such as formative assessments [26]. Conversely, an MCQ that exhibits two or more IWFs is classified as unacceptable for use. However, instructors might prioritize avoiding specific IWFs based on their use cases to align best with their learning objectives.

**Table 1.** The 19 Item-Writing Flaw rubric criteria used in this study.

| Item-Writing Flaw | An Item Is Flawed If... |
|---|---|
| Longest Option Correct | The correct option is longer and includes more detailed information than the other distractors, as this clues students to this option |
| Ambiguous Information | The question text or any of the options are written in an unclear way that includes ambiguous language |
| Implausible Distractors | Any included distractors are implausible, as good items depend on having effective distractors |
| True or False | The options are a series of true/false statements |
| Absolute Terms | It contains he use of absolute terms (e.g. never, always, all) in the question text or options |
| Complex or K-type | It contains a range of correct responses that ask students to select from a number of possible combinations of the responses |
| Negatively Worded | The question text is negatively worded, as it is less likely to measure important learning outcomes and can confuse students |
| Convergence Cues | Convergence cues are present in the options, where there are different combinations of multiple components to the answer |
| Lost Sequence | The options are not arranged in chronological or numerical order |
| Unfocused Stem | The stem is not a clear and focused question that can be understood and answered without looking at the options |

| None of the Above | One of the options is "none of the above", as it only really measures students ability to detect incorrect answers |
|---|---|
| Word Repeats | The question text and correct response contain words only repeated between the two |
| More Than One Correct | There is not a single best-answer, as there should be only one answer |
| Logical Cues | It contains clues in the stem and the correct option that can help the test-wise student to identify the correct option |
| All of the Above | One of the options is "all of the above", as students can guess correct responses based on partial information |
| Fill in the Blank | The question text omits words in the middle of the stem that students must insert from the options provided |
| Vague Terms | It uses vague terms (e.g. frequently, occasionally) in the options, as there is seldom agreement on their actual meaning |
| Grammatical Cues | All options are not grammatically consistent with the stem, as they should be parallel in style and form |
| Gratuitous Information | It contains unnecessary information in the stem that is not required to answer the question |

### 3.2 Technical Overview of SAQUET

Previous efforts to automate the application of the IWF rubric have explored two main strategies, using either a rule-based approach or the well-known GPT-4 model [19]. The rule-based approach demonstrated superior performance to the GPT-4-based method for most criteria across all domains used in the previous study. Building upon these findings, this current work enhances the rule-based methodology by integrating advanced methods and incorporating selective GPT-4 interventions. One of our primary objectives was not only to improve the quality of criteria classifications, but also to preserve the tool's ability to be applied across various domains, ensuring scalability and rapid processing for a large volume of MCQs. The automatic detection of the 19 IWF criteria outlined in Table 1 falls into three distinct categories: text-matching techniques, NLP-based information extraction, and enhancements provided by GPT-4.

The first category includes eight criteria: *None of the Above, All of the Above, Fill-In-The-Blank, True or False, Longest Answer Correct, Negative Worded, Lost Sequence,* and *Vague terms*. Given the nature of these criteria, foundational programming techniques like string matching are primarily used for identification. However, to enhance accuracy we implemented several modifications, such as adjusting threshold parameters, incorporating checks for various question formats, expanding the list of keywords for matching, and lemmatizing the text to normalize word forms. For example, the *True or False* criteria underwent significant alterations to accommodate Yes/No questions. The *Fill-In-The-Blank* criteria required adjustments to avoid misclassification of Computer Science MCQs, which often use the underscore character. Improvements like refined pattern matching were applied to the *Lost Sequence* criteria, enabling the detection of cases not identified in the initial dataset.

The second category encompasses five criteria: *Implausible Distractors*, *Word Repeats, Logical Cues, Ambiguous or Unclear,* and *Grammatical Cues*. These criteria are addressed using foundational NLP techniques, including word embeddings, Named Entity Recognition (NER), and Transformer models like RoBERTa [21]. NER plays a pivotal role in analyzing *Word Repeats, Logical Cues*, and *Grammatical Cues* by allowing us to identify and compare nouns and verbs used in the MCQ. This approach enhances our ability to detect grammatical consistency, identify repeated words, and recognize synonyms. For tackling *Ambiguous Information* and *Implausible Distractors*, our attempts to incorporate GPT-4 faced challenges, as its outputs were often excessively critical, leading to a high rate of misclassifications. To address this, we instead integrated additional linguistic metrics, such as query well-formedness scores [8], and leveraged updated word embeddings to refine the evaluation.

The final category includes six criteria: *Absolute Terms, More Than One Correct, Complex or K-Type, Gratuitous Information, Unfocused Stem,* and *Convergence Cues*. This category utilizes NLP techniques similar to the previous ones, enhanced by the integration of GPT-4 API calls for additional verification. For example, simple word matching was insufficient for the *Absolute Terms* criteria, as the context in which terms like "impossible" are used needs further analysis by GPT-4 to determine their impact on answer validity. Modifications were applied to the *Convergence Cues* and *Complex or K-Type* criteria, incorporating GPT-4 for final verification check to improve accuracy. The criteria *Unfocused Stem* and *Gratuitous Information*, both of which involve lexical richness [11], benefited from GPT-4 interventions, significantly reducing false positives detected in pilot tests by better evaluating question stems for learner comprehension. Finally, the *More Than One Correct* criteria was enhanced to not only attempt at answering questions but also to discern whether a question allows for multiple correct responses or is a select-all-that-apply type. We have open-sourced the code and datasets used in this work[i].

### 3.3   Datasets

We utilized two datasets of MCQs previously tagged with the IWF criteria to evaluate SAQUET. The first dataset, derived from [5], encompasses MCQs in Computer Science, Humanities, and Healthcare, sourced from prominent MOOC platforms, such as Coursera and edX. The second dataset, from [19], contains student-generated MCQs from Chemistry and Statistics courses. Both datasets contained MCQs with two to five answer choices each. Additionally, both datasets were evaluated by two human experts, with past studies reporting high inter-rater reliability via Kappa scores. Due to IRB permissions and formatting challenges, not all questions from these initial datasets were included in our present study. Additionally, we made minor corrections to address errors in the datasets, such as mislabeled criteria. For example, one adjustment involved reevaluating Computer Science, Humanities, and Healthcare questions to ensure True/False questions were not mistakenly flagged under the *Longest Option Correct* criteria, particularly when "False" was the correct answer.

For developing SAQUET, we initially used a subset of 25 questions, 5 from each domain, which were not included in the final evaluation dataset. Our final dataset comprised 271 MCQs across the five domains, all tagged with the 19 IWF criteria, offering a varied pool of questions for analysis. This contrasts with previous IWF research, which often focuses on a single domain [23, 26].

### 3.4   Evaluation

To evaluate the effectiveness of commonly employed automated techniques for assessing question quality, we applied five popular linguistic quality metrics to the 271

MCQs in our dataset: perplexity, diversity, grammatical error, complexity, and answerability. Perplexity scores were generated using a GPT-2 language model, aligning with methodologies from recent research [28]. We measured diversity through the Distinct-3 score, which quantifies the average number of unique 3-grams per MCQ [13]. Grammatical errors were identified using the widely recognized Python Language Tool [20], tallying the grammatical inaccuracies in each question as done in prior research [24]. For complexity assessment, we adopted Bloom's Revised Taxonomy, assigning each MCQ a level from 0 (lowest, 'remember') to 5 (highest, 'create'), which serves as a common indicator of complexity and difficulty [11, 17]. A highly precise classifier was employed to automatically determine the Bloom's level for each question [6]. Answerability was evaluated using GPT-4, employing the strategy of the Prompting-based Metric on ANswerability (PMAN) approach [27]. This involved following the strategy of crafting specific prompts that instructed GPT-4 to choose an answer for each MCQ.

For the evaluation of SAQUET, we referenced gold standard human evaluations for our dataset. The overall match rate between our method and the human evaluations is calculated to reflect the general accuracy of our tool in classifying MCQs according to the IWF criteria. To tackle this multi-label classification challenge, we use the exact match ratio, necessitating correct identification of all labels for a match, and Hamming Loss, which calculates the average proportion of incorrect labels, offering detailed insights into our classification's accuracy on a holistic level [9]. We further assess performance using the F1 score of each criteria, which balances precision (the accuracy of positive predictions) and recall (the completeness of positive predictions) [4]. A high F1 score indicates both high precision and high recall, signifying effective identification of an IWF without excessive false positives or negatives. The micro-averaged F1 score aggregates outcomes across all criteria, offering a consolidated view of performance for the entire dataset [14]. Analysis is conducted not just on the aggregate dataset, but also segmented by domain. This allows us to identify domain-specific performance variations and areas for refinement. Where possible, we compare our results with metrics reported in prior studies using similar datasets and evaluation metrics, providing context for SAQUET's performance [19, 5].

## 4    Results

### 4.1    Limitations of Traditional Metrics in Evaluating Educational MCQs

For each of the five domains, we categorized the MCQs into two groups: one group includes MCQs with zero or one IWF and the other comprises MCQs with two or more. This classification helps differentiate between questions that are considered acceptable (zero or one IWF) and those deemed unacceptable (two or more IWF), thereby allowing for a more precise analysis given the constraints of our dataset in accordance with previous research [19, 26]. We then assessed these questions using five linguistic quality evaluation metrics, as detailed in Table 2. Our analysis revealed that, across all metrics, the performance of MCQs in each domain either matched or exceeded ones found in recent research. For comparison, [4] reported that human generated MCQs, based on Wikipedia articles and science textbooks, had average perplexity scores of 18 to 84 and diversity scores between .78 and .82. Similarly, [24] determined that the average answerability score for human generated MCQs, on the topic of middle and high school reading comprehension, was .726.

**Table 2.** Comparison of five common evaluation metrics for question quality across five domains, categorized by IWF Count. A circumflex (^) denotes a superior score achieved by questions with a higher IWF count in each metric.

| Domain | IWF | N | Perplexity ↓ | Diversity ↑ | Grammatical Error ↓ | Cognitive Complexity ↑ | Answerability ↑ |
|---|---|---|---|---|---|---|---|
| Chemistry | 0-1 | 35 | 47.65 | 0.961 | 0.400 | 0.057 | 0.743 |
|  | 2+ | 15 | 57.46 | 0.962(^) | 0.333(^) | 0.133(^) | 0.733 |
| Statistics | 0-1 | 32 | 46.02 | 0.928 | 0.375 | 0.719 | 0.531 |
|  | 2+ | 18 | 27.51(^) | 0.888 | 0.444 | 1.333(^) | 0.611(^) |
| Computer Science | 0-1 | 62 | 30.73 | 0.927 | 2.129 | 1.145 | 0.806 |
|  | 2+ | 38 | 41.56 | 0.917 | 3.605 | 1.500(^) | 0.605 |
| Humanities | 0-1 | 18 | 47.64 | 0.955 | 0.375 | 1.313 | 0.875 |
|  | 2+ | 6 | 28.24(^) | 0.939 | 0.375 | 1.250 | 1.000(^) |
| Healthcare | 0-1 | 25 | 30.25 | 0.955 | 0.400 | 1.200 | 0.960 |
|  | 2+ | 22 | 27.72(^) | 0.957(^) | 0.182(^) | 1.682(^) | 0.909 |

Our analysis revealed that student-generated questions in the Chemistry and Statistics domains had relatively high perplexity scores, but in Statistics, questions with 2+ IWFs exhibited a lower perplexity. The diversity metric revealed a ceiling effect, where variations are minimal across different question sets from all domains. High diversity scores are expected, as the MCQs were sourced from diverse origins and authors, such as MOOCs or digital textbooks. The impact of IWFs on a question's answerability varied, where in some cases the presence of IWFs did not reduce, and might have even enhanced, the likelihood of the LLM to correctly answer the questions.

Grammatical errors were relatively low across all domains except Computer Science, where the code syntax posed unique challenges for this criteria, contributing to higher error rates [6]. Interestingly, in both Chemistry and Healthcare, questions with more IWFs (2+) showed a lower average number of grammatical errors, suggesting a nuanced relationship between IWF count and grammatical precision. Initially we expected questions with fewer IWF would have fewer grammatical mistakes, but those may have been overlooked by the human evaluators. Cognitive complexity, measured by Bloom's Revised Taxonomy levels, was also generally higher for questions with 2+ IWFs across all domains except for Humanities, where the difference was marginal, indicating these questions with more flaws tend to engage higher-order cognitive skills.

These findings demonstrate the potential for commonly used metrics to paint an overly optimistic picture of question quality. Even questions with multiple flaws can score well on perplexity, diversity, and grammatical precision, suggesting they are well-crafted and clear. However, this can be misleading, as these metrics may not capture deeper issues such as false information, incorrect assumptions, or inaccuracies in content. For example, Figure 1 shows a question that achieved an acceptable evaluation across all five metrics, yet it is clearly a poorly student generated question that contains three IWFs: *implausible distractors, logical cues,* and *grammatical cues.*

What is protons?  
    A) positively charged particles

**Perplexity**: 27.56  
**Diversity**: 1.0

B) sum the number of protons and neutrons  
C) negatively charged subatomic particles  
D) he discovered the charge of electron  

**Grammatical Error**: 1  
**Complexity**: 0 (remember)  
**Answerability**: 1

**Fig. 1.** A student generated MCQ from the Chemistry dataset consisting of three IWFs on the left, with the associated linguistic quality evaluation metrics on the right.

### 4.2   Performance of Automated IWF Classification Across Domains

The 19 IWF criteria were automatically applied to all 271 MCQs for a total of 5,149 classifications. While the overall accuracy is slightly skewed due to most of the questions containing a few flaws and thus being classified as 0 for a given criteria, the total accuracy was 94.13%, which treats each criteria classification individually. We achieved an exact match ratio of 38%, which indicates that 103 of the questions were evaluated the same across all 19 criteria between SAQUET and the different human evaluators. The Hamming Loss was 5.9%, indicating a small amount of misclassification regarding the flaws. While we only used half of the data from [19] consisting of 100 MCQs, it is our closest comparable. As such, compared to their leading rule-based method, we achieved a 3.26% overall classification accuracy improvement, a 13% higher exact match ratio, and 3.1% lower Hamming Loss.

On average, SAQUET (*M*=1.75, *SD*=1.26) was more likely to classify a MCQ as having more IWFs compared to the human evaluators (*M*=1.31, *SD*=1.11). The most IWFs assigned to a single question by both was 5. In Table 3, we present the IWF classifications from the human evaluators compared to SAQUET for all five domains.

**Table 3.** The number of identified flaws (N) and F1 performance scores for human evaluations (Hum) versus SAQUET (SAQ) across the five domains. A dash (-) signifies the absence of a flaw in a domain as determined by human evaluation, precluding F1 score calculation.

| Item-Writing Flaws | | Chemistry (50) | | Statistics (50) | | Computer Science (100) | | Humanities (24) | | Healthcare (47) | |
|---|---|---|---|---|---|---|---|---|---|---|---|
| | | Hum | SAQ | Hum | SAQ | Hum | SAQ | Hum | SAQ | Hum | SAQ |
| Longest Option Correct | N | 5 | 8 | 3 | 7 | 27 | 27 | 8 | 8 | 16 | 15 |
| | F1 | 0.77 | | 0.60 | | 0.96 | | 1.00 | | 0.97 | |
| Ambiguous Information | N | 12 | 12 | 14 | 18 | 12 | 21 | 0 | 2 | 2 | 0 |
| | F1 | 0.58 | | 0.50 | | 0.24 | | 0.00 | | 0.00 | |
| Implausible Distractors | N | 9 | 8 | 8 | 6 | 3 | 15 | 3 | 7 | 8 | 3 |
| | F1 | 0.24 | | 0.86 | | 0.33 | | 0.20 | | 0.00 | |
| True or False | N | 2 | 2 | 1 | 0 | 9 | 10 | 4 | 4 | 11 | 11 |
| | F1 | 1.00 | | 0.00 | | 0.95 | | 1.00 | | 1.00 | |
| Absolute Terms | N | 2 | 1 | 0 | 1 | 9 | 6 | 9 | 9 | 5 | 4 |
| | F1 | 0.67 | | 0.00 | | 0.40 | | 0.89 | | 0.44 | |
| Complex or K-type | N | 2 | 4 | 4 | 8 | 15 | 12 | 0 | 1 | 4 | 5 |
| | F1 | 0.67 | | 0.67 | | 0.81 | | 0.00 | | 0.89 | |
| Negatively Worded | N | 0 | 0 | 2 | 4 | 10 | 14 | 0 | 1 | 11 | 11 |
| | F1 | - | | 0.67 | | 0.83 | | 0.00 | | 0.91 | |
| Convergence Cues | N | 2 | 3 | 9 | 7 | 7 | 11 | 0 | 0 | 1 | 4 |
| | F1 | 0.00 | | 0.63 | | 0.44 | | - | | 0.00 | |
| Lost Sequence | N | 3 | 3 | 14 | 15 | 2 | 2 | 0 | 0 | 0 | 0 |
| | F1 | 1.00 | | 0.97 | | 0.50 | | - | | - | |

| | | | | | | | | | | |
|---|---|---|---|---|---|---|---|---|---|---|
| Unfocused Stem | N<br>F1 | 0<br>0.00 | 1 | 8<br>0.89 | 10 | 8<br>0.62 | 5 | 0<br>- | 0 | 0<br>- | 0 |
| None of the Above | N<br>F1 | 6<br>0.91 | 5 | 1<br>1.00 | 1 | 6<br>1.00 | 6 | 0<br>- | 0 | 0<br>- | 0 |
| Word Repeats | N<br>F1 | 1<br>1.00 | 1 | 1<br>1.00 | 1 | 7<br>0.56 | 11 | 0<br>- | 0 | 4<br>0.53 | 11 |
| More Than One Correct | N<br>F1 | 0<br>0.00 | 2 | 0<br>0.00 | 11 | 8<br>0.38 | 24 | 3<br>0.46 | 10 | 1<br>0.11 | 17 |
| Logical Cues | N<br>F1 | 4<br>0.29 | 3 | 2<br>0.67 | 1 | 2<br>0.00 | 8 | 0<br>- | 0 | 0<br>0.00 | 1 |
| All of the Above | N<br>F1 | 1<br>1.00 | 1 | 1<br>1.00 | 1 | 2<br>1.00 | 2 | 0<br>- | 0 | 2<br>0.80 | 3 |
| Fill in the Blank | N<br>F1 | 2<br>1.00 | 2 | 0<br>- | 0 | 2<br>1.00 | 2 | 0<br>- | 0 | 2<br>1.00 | 2 |
| Vague Terms | N<br>F1 | 0<br>- | 0 | 0<br>0.00 | 1 | 3<br>0.80 | 2 | 0<br>0.00 | 1 | 3<br>1.00 | 3 |
| Grammatical Cues | N<br>F1 | 2<br>0.67 | 1 | 3<br>0.00 | 1 | 0<br>0.00 | 1 | 0<br>0.00 | 1 | 0<br>0.00 | 1 |
| Gratuitous Information | N<br>F1 | 0<br>0.00 | 2 | 3<br>0.50 | 5 | 0<br>0.00 | 3 | 0<br>0.00 | 2 | 0<br>- | 0 |
| Micro-Averaged F1 | | 0.59 | | 0.65 | | 0.62 | | 0.66 | | 0.67 | |
| IWF totals | | 53 | 59 | 74 | 98 | 132 | 182 | 27 | 46 | 70 | 91 |

The F1 scores reveal the effectiveness of SAQUET across the five domains for each criterion. Compared to the rule-based implementation in [19], our approach improved the F1 score across multiple criteria for Chemistry and Statistics questions. Performance on the *None of the Above* criteria was notably strong, as reflected by high F1 scores, indicating precise classification with minimal misclassifications. Other criteria, such as *More Than One Correct*, showed subpar performance across all domains, with frequent incorrect classifications and often overestimating its presence. The micro-averaged F1 scores provide a consolidated view of SAQUET's accuracy across all 19 criteria and allow for a domain-wise comparison of classification efficacy.

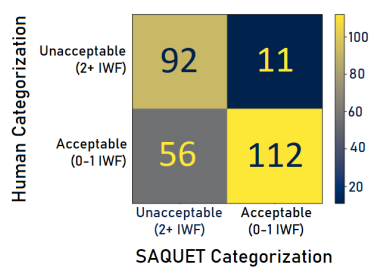

**Fig. 2.** A confusion matrix for the categorization of questions as acceptable or unacceptable based on their IWF by the human evaluation and SAQUET.

Taking the categorization of all MCQs as acceptable (zero or one IWF) or unacceptable (two or more IWF), we compared the SAQUET's classifications with those made by human evaluators. This comparison aimed to see if the overall categorization matched, despite potential misclassifications of specific IWF criteria. Figure 2 presents a confusion matrix for this acceptability classification, indicating human evaluators deemed 168 questions acceptable and 103 questions unacceptable. SAQUET matched 204 of these MCQ categorizations, with 112 classified as acceptable and 92 as unacceptable, achieving a 75.3% match rate with human evaluations.

## 5    Discussion

Our results demonstrate that traditional metrics used for assessing the quality of questions, especially multiple-choice, might not adequately reflect their true quality. We observed that questions with various errors, indicated by Item-Writing Flaws, which could either simplify the answering process for students or lead to confusion, often receive high scores from commonly used linguistic quality metrics. To address this gap, we introduced SAQUET, a method designed to capture these more complex aspects of question quality while remaining automated and scalable. By benchmarking against human expert evaluations, we show that SAQUET has the potential to provide a more precise and detailed assessment of question quality compared to these linguistic quality metrics. Furthermore, our contribution to the field of assessment quality evaluation research extends to making both SAQUET and our comprehensive dataset publicly available[i].

Recent efforts in NLP have aimed to shift away from traditional readability metrics like BLEU, METEOR, or ROGUE when evaluating the quality of MCQs, yet these metrics continue to be employed in recent works [2, 4, 21]. In our study, we explored alternative linguistic quality metrics (perplexity, diversity, grammar, complexity, answerability) that are also commonly used and offer a different approach to question evaluation, particularly in response to the inadequacies of previous readability metrics [12, 16, 27]. Our findings reveal that even questions with obvious flaws can be evaluated as higher quality according to these metrics. This discrepancy may still hold for machine generated questions from older models, but the improved linguistic capabilities of recent LLMs mean that more machine generated questions are likely to be deemed high quality by these standards. Recent studies have pointed out that despite the grammatical correctness of LLM outputs, the MCQs generated can suffer from issues like implausible distractors or vague wording [7, 21].

SAQUET has the advantage of operating without training data, addressing the significant challenge of sourcing IWF-tagged question datasets. Although research utilizing the IWF rubric is widespread, access to such datasets is often restricted. Importantly, SAQUET's application extends beyond assessing newly crafted questions; it is equally effective in evaluating existing question sets and machine- or human generated questions alike. This capability allows educators to pinpoint and address flaws in current questions they might be using, potentially adjusting or replacing them to suit their needs. In this study, we achieved an exact match ratio of 38% in a complex multi-label classification task with 19 binary labels, which serves as a strong baseline for future research and evaluation. When compared to human evaluations, SAQUET showed a propensity to identify IWFs more frequently. We prefer this stricter approach of identifying MCQ flaws while prioritizing false positives over false negatives, thereby ensuring only the highest quality questions are utilized for educational purposes.

For the criteria based primarily on text matching, such as *True or False*, *All of the Above*, or *Longest Option Correct*, one might intuitively expect perfect accuracy. However, our findings indicate that these criteria can manifest in nuanced forms, demonstrating the importance of datasets that capture a broader spectrum of these errors. For instance, True/False MCQs might also appear as Yes/No choices or contain explanation text that follows the option, complicating their identification. Similarly, interpretations of what constitutions *Longest Option Correct* can vary among human evaluators, as it did in our study. In Chemistry and Statistics this flaw was applied to questions if the second-longest option was not nearly as long (at least 80%) as the longest. In contrast, for Computer Science, Humanities, and Healthcare, a stricter interpretation was applied that flagged any question where the correct answer exceeded others in length by even a single character.

Other flaws like *More than one Correct*, which relied heavily on GPT-4, presented significant challenges, notably impacting the overall exact match ratio. This flaw saw a misclassification for 50 out of 271 questions (18.5%), making it the most problematic. The challenge arose from GPT-4's difficulty in reliably identifying the correct answer for an MCQ, frequently failing to determine if a single correct option exists. However, this limitation is not inherently negative, as it does not imply the question is flawed, just that the LLM has the inability to solve it [18, 27]. This highlights the ongoing challenge of accurately evaluating complex question criteria and the limitations of current AI in navigating such nuances, further emphasizing the need for refined and open approaches along with diverse datasets in the evaluation process.

## 6      Limitations and Future Work

In our study, we introduced SAQUET, an automated method for evaluating questions, employing multiple criteria that leverage LLMs like GPT-4. While outperforming traditional automated MCQ evaluation metrics, this approach comes with inherent limitations, including the black box nature of LLMs, their potential for unanticipated changes, and the risk of bias in their outputs. To mitigate these issues and enhance this work's reliability and cost-effectiveness, we utilized a specific version of GPT-4 through the `gpt-4-0125-preview`[ii] API. This approach aimed to standardize the evaluation process and ensure reproducibility by generating consistent outputs from predefined prompts. We further supported transparency and reproducibility by open-sourcing our code[i]. Expanding our dataset to include a greater number and diversity of questions across additional domains would likely reveal further limitations and areas for improvement in our current evaluation criteria.

For future work, we aim to enhance the evaluation techniques for the 19 IWF criteria, with a particular focus on those that currently show weaker performance. Acquiring additional datasets of MCQs annotated with IWFs will be crucial in validating and demonstrating the effectiveness of our method. We encourage educators, researchers, and practitioners to engage with our work, offering their insights and improvements to refine the criteria further, as we have done. Such collaboration would contribute to developing a more educationally robust metric enriched by collective expertise. As LLMs advance, we anticipate that our methodology will too, achieving greater accuracy for certain criteria and providing detailed feedback on how to correct identified flaws.

An Automatic Question Usability Evaluation Toolkit    13## 7  Conclusion

In this study, we highlight the limitations of current metrics for assessing question quality, particularly their oversight of deeper question attributes beyond mere surface characteristics. Through analyzing a dataset of MCQs spanning five varied domains, we illustrate that these prevalent linguistic quality metrics fall short in effectively differentiating between flawed and flawless questions. This gap demonstrates the need for a novel metric capable of comprehensive question quality evaluation. In response, we refined an alternative evaluation method that retains both automation and scalability by assessing MCQs against a detailed 19-criteria Item-Writing Flaws rubric. Upon validating this method to our dataset, we demonstrated its effectiveness across various domains and identified the criteria that were most and least effective. Our findings reveal the potential to significantly enhance question quality assessment, paving the way for more accurate and educationally valuable evaluations.

## References

1.  Azevedo, J.M., Oliveira, E.P., Beites, P.D.: Using learning analytics to evaluate the quality of multiple-choice questions: A perspective with classical test theory and item response theory. Int. J. Inf. Learn. Technol. 36, 4, 322–341 (2019).
2.  Bhowmick, A.K., Jagmohan, A., Vempaty, A., Dey, P., Hall, L., Hartman, J., Kokku, R., Maheshwari, H.: Automating Question Generation From Educational Text. In: Artificial Intelligence XL. pp. 437–450 Springer Nature Switzerland, Cham (2023).
3.  Bitew, S.K., Deleu, J., Develder, C., Demeester, T.: Distractor generation for multiple-choice questions with predictive prompting and large language models. In: RKDE2023, the 1st International Tutorial and Workshop on Responsible Knowledge Discovery in Education Side event at ECML-PKDD. (2023).
4.  Bulathwela, S., Muse, H., Yilmaz, E.: Scalable Educational Question Generation with Pre-trained Language Models. In: Wang, N., Rebolledo-Mendez, G., Matsuda, N., Santos, O.C., and Dimitrova, V. (eds.) Artificial Intelligence in Education. pp. 327–339 Springer Nature Switzerland, Cham (2023). https://doi.org/10.1007/978-3-031-36272-9_27.
5.  Costello, E., Holland, J.C., Kirwan, C.: Evaluation of MCQs from MOOCs for common item writing flaws. BMC Res. (2018). https://doi.org/10.1186/s13104-018-3959-4.
6.  Doughty, J. et al.: A Comparative Study of AI-Generated (GPT-4) and Human-crafted MCQs in Programming Education. In: Proceedings of the 26th Australasian Computing Education Conference. pp. 114–123 ACM, Sydney NSW Australia (2024).
7.  Elkins, S., Kochmar, E., Cheung, J.C.K., Serban, I.: How Teachers Can Use Large Language Models and Bloom's Taxonomy to Create Educational Quizzes. In: Proceedings of the AAAI Conference on Artificial Intelligence. (2024).
8.  Faruqui, M., Das, D.: Identifying Well-formed Natural Language Questions. In: Proceedings of the 2018 Conference on Empirical Methods in Natural Language Processing. pp. 798–803 (2018).
9.  Ganda, D., Buch, R.: A survey on multi label classification. Recent Trends Program. Lang. 5, 1, 19–23 (2018).
10. Kasneci, E., Seßler, K., Küchemann, S., Bannert, M., Dementieva, D., Fischer, F., Gasser, U., Groh, G., Günnemann, S., Hüllermeier, E.: ChatGPT for good? On opportunities and challenges of large language models for education. Learn. Individ. Differ. 102274 (2023).
11. Kurdi, G., Leo, J., Parsia, B., Sattler, U., Al-Emari, S.: A systematic review of automatic question generation for educational purposes. Int. J. Artif. Intell. Educ. 30, (2020).

---

[i] https://github.com/StevenJamesMoore/AIED24

[ii] https://platform.openai.com/docs/models/gpt-4-and-gpt-4-turbo